%% file: output.tex
\documentclass[runningheads]{llncs}
\input{math_commands.tex}

\usepackage{eccv}

\usepackage{eccvabbrv}
\usepackage{fontawesome5}
\usepackage{multirow,makecell,graphicx}
\usepackage{wrapfig}
\usepackage{enumitem}
\usepackage{orcidlink}
\usepackage{hyperref}
\usepackage{subcaption}
\usepackage{graphicx}
\usepackage{booktabs}
\usepackage{url}
\usepackage[accsupp]{axessibility}  
\newtheorem{assumption}{Assumption}

\usepackage[capitalize]{cleveref}
\crefname{section}{Sec.}{Secs.}
\Crefname{section}{Section}{Sections}
\Crefname{table}{Table}{Tables}
\crefname{table}{Tab.}{Tabs.}
\crefname{equation}{Eq.}{Eqs.}
\crefname{algorithm}{Algorithm}{Algorithm}
\crefname{assumption}{Assumption }{Assumption }

\begin{document}

\title{LPViT: Low-Power Semi-structured Pruning for \\Vision Transformers} 

\titlerunning{Abbreviated paper title}

\author{Kaixin Xu\inst{1,2}$^\ast$\orcidlink{0000-0002-7222-2628} \and
Zhe Wang\inst{1,2}$^\ast$ \and
Chunyun Chen\inst{2} \and
Xue Geng\inst{1} \and
Jie Lin\inst{1} \and
Mohamed M. Sabry Aly\inst{2} \and
Xulei Yang\inst{1} \and
Min Wu\inst{1} \textsuperscript{\faEnvelope[regular]} \and
Xiaoli Li\inst{1} \and
Weisi Lin\inst{2} }

\authorrunning{K.~Xu et al.}

\institute{Institute for Infocomm Research (I$^2$R), Agency for Science, Technology and Research (A*STAR), 1 Fusionopolis Way, 138632, Singapore 
\email{\{xuk,wang\_zhe,geng\_xue,yang\_xulei,wumin,xlli\}@i2r.a-star.edu.sg}
\email{jie.dellinger@gmail.com} \and
College of Computing and Data Science (CCDS), Nanyang Technological University (NTU), Singapore \\
\email{chunyun001@e.ntu.edu.sg,msabry@ntu.edu.sg,wslin@ntu.edu.sg}
}

\maketitle

\let\thefootnote\relax\footnotetext{$^\ast$ Equal contribution. \faEnvelope[regular] Corresponding author.}

\begin{abstract}
  Vision transformers (ViTs) have emerged as a promising alternative to convolutional neural networks (CNNs) for various image analysis tasks, offering comparable or superior performance. However, one significant drawback of ViTs is their resource-intensive nature, leading to increased memory footprint, computation complexity, and power consumption. To democratize this high-performance technology and make it more environmentally friendly, it is essential to compress ViT models, reducing their resource requirements while maintaining high performance.
In this paper, we introduce a new block-structured pruning to address the resource-intensive issue for ViTs, offering a balanced trade-off between accuracy and hardware acceleration. Unlike unstructured pruning or channel-wise structured pruning, block pruning leverages the block-wise structure of linear layers, resulting in more efficient matrix multiplications. To optimize this pruning scheme, our paper proposes a novel hardware-aware learning objective that simultaneously maximizes speedup and minimizes power consumption during inference, tailored to the block sparsity structure. This objective eliminates the need for empirical look-up tables and focuses solely on reducing parametrized layer connections.
Moreover, our paper provides a lightweight algorithm to achieve post-training pruning for ViTs, utilizing second-order Taylor approximation and empirical optimization to solve the proposed hardware-aware objective. Extensive experiments on ImageNet are conducted across various ViT architectures, including DeiT-B and DeiT-S, demonstrating competitive performance with other pruning methods and achieving a remarkable balance between accuracy preservation and power savings. Especially, we achieve up to $\mathbf{3.93}\times$ and $\mathbf{1.79}\times$ speedups on dedicated hardware and GPUs respectively for DeiT-B, and also observe an inference power reduction by $\mathbf{1.4}\times$ on real-world GPUs. 
Code has been released to \url{https://github.com/Akimoto-Cris/LPViT}.
\end{abstract}

\section{Introductions}

Recently, vision transformers (ViTs) have been an emerging string of research that greatly challenges the prevailing CNNs with on-par or even superior performance on various image analysis and understanding tasks such as classification~\cite{dosovitskiy2020image,Cordonnier2020On,deit,tnt,he2022masked}, object detection~\cite{carion2020end,zhu2021deformable,amini2021t6d}, semantic segmentation~\cite{chen2021pre,liu2021swin}, etc., but completely without the convolution mechanism seen in the CNNs. 
\begin{wrapfigure}{r}{0.35\textwidth}
    \centering
    \includegraphics[width=.35\textwidth]{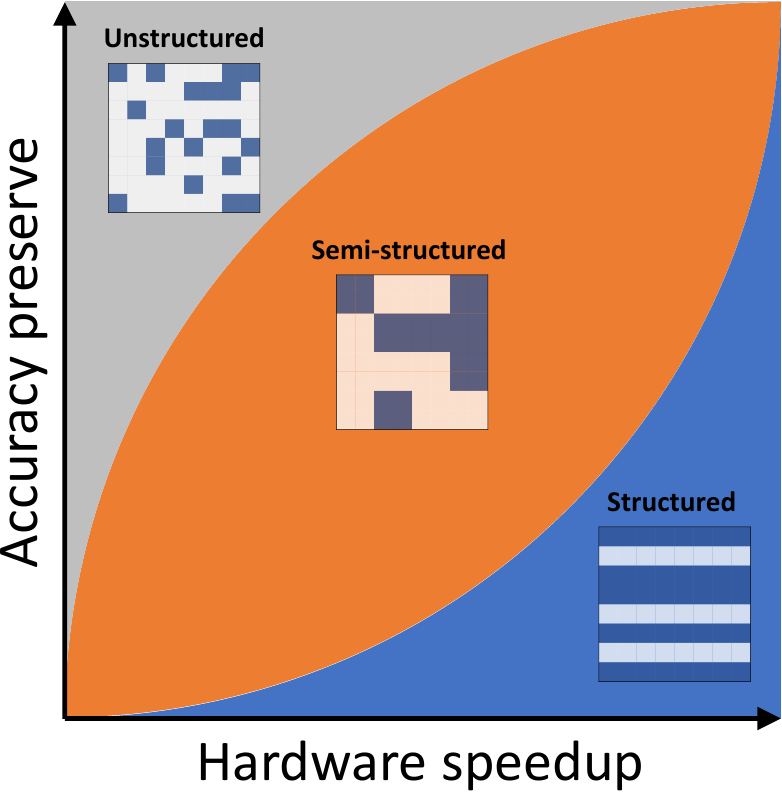}
    \caption{Trade-offs of different sparsity schemes in terms of model accuracy and hardware acceleration.}
    \label{fig:block}
    \vspace{-15pt}
\end{wrapfigure}
Despite the success in the task performances, as pointed out by~\cite{uvc}, one major drawback of the ViTs architecture is that the ViTs are much less resource-efficient than CNNs in terms of memory footprint, computation complexity and the eventual power consumption. 
To make the high-performance ViTs more environmental friendly and democratize the technology, it is necessary to compress the ViTs models and cut down the power consumption, so that they could be accessed by low-end computation devices with equal or comparable model performance.

Among different bifurcations of neural network compression, network pruning is an effective method that has shown success on CNNs, which prunes out redundant neurons or rules out computations in the networks.
Previously on CNNs, some{~\cite{han2015deep,han2015learning,zhu2018prune,lamp,morcos2019one,lin2020hrank,wang2022rdo,xu2023efficient} attempted \textit{unstructured pruning} to the models which removes individual neurons from the layer weights; while others\cite{luo2017thinet,shen2022structural} used \textit{structured pruning} which removes channel-wise neurons. Comparing to unstructured pruning, the latter structured scheme has high data locality hence is more hardware-friendly~\cite{buluc2008challenges} as it is easier to achieve accelerated computation by simply removing entire rows or columns in the weight matrices, it cause severer accuracy degradation due to the coarser pruning granularity making it a much more challenging pruning scheme. 


Nevertheless, for transformer architectures consisting of mostly linear layers (matrix multiplication), block structured (semi-structured) pruning is a better trade off between accuracy and hardware acceleration, since the GEMM performs matrix multiplication in a block-by-block manner. Hence multiplication with block sparse matrices can achieve more speedup than unstructured ones under the same pruning ratio while still maintaining high accuracy. A summarized qualitative comparison among pruning schemes is listed in Fig.~\ref{fig:block}. 
Prior arts~\cite{mao2021tprune,lagunas2021block} in NLP domain validated the block structured pruning on language models (BERT~\cite{devlin2018bert}, MobileBERT~\cite{sun2020mobilebert}, etc.), achieving more than $2\times$ speedup with negligible performance drop. However, the other parts of their pruning scheme is rather out-dated, \emph{e.g.} vanilla pruning criterion. Similar attempts are still scarce on ViTs for various vision tasks. 


In this work, we propose a novel block-structured pruning approach for ViTs to prune the parameters in a block-based manner to achieve better trade-off between accuracy and efficiency. We formulate the learning objective in a way that simultaneously maintains the accuracy of the pruned model and minimizes the number of the computational operations. A hardware-aware constraint is incorporated into the objective to boost the speedup and lower power consumption during inference stage.
Moreover, we present a fast optimization method to solve the objective function by utilizing second-order Taylor approximation. 
After equivalent reformulation, such we are able to solve the objective very efficiently (quadratic to cubic complexity for empirical data collection against network size and linear time complexity for equation solving).
To the best of our knowledge, this is the first paper that introduces the block-structured pruning scheme and present a hardware-aware post-training pruning approach for ViTs. The main contributions are summarized as below:

\begin{itemize}
    \item We systematically formulate an optimal hardware-aware pruning objective for ViTs models under the block-structured pruning scheme, which directly optimizes both model accuracy and power consumption at the same time. The power consumption is fully estimated without constructing any empirical look-up tables (LUTs), making it a light-weight approach and does not require additional overheads for optimization. The proposed pruning scheme solely relies on reducing parametrized layer connections without manipulating skip configurations and token pruning.
    \item We then provide an efficient solution for the proposed hardware-aware objective function by second-order taylor approximation and present an empirical optimization method with only linear time complexity. 
    With the weights being pruned, less parameters are required to read/write from off-chip memory to on-chip memory leading to memory reduction and significant power reduction.
    
    \item Extensive experiments demonstrate the effectiveness of our approach. Results on various deep ViTs architectures, including DeiT-B, DeiT-S, Swin-T, Swin-B, etc. show that our approach noticeably outperforms the state-of-the-arts while with up to $\mathbf{3.93}\times$ and $\mathbf{1.79}\times$ speedups on dedicated hardware and GPUs respectively for DeiT-B. Inference power reduction by $\mathbf{1.4}\times$ is also witnessed on real-world GPUs.
    
\end{itemize}

\section{Related Works}
\subsection{Vision Transformers (ViTs)}
Following the success of self-attention based transformer architecture in natural language processing~\cite{vaswani2017attention}, transformer based vision models have also been marching in image domain and being strong competitors against traditional CNNs in various scenes like object detection~\cite{carion2020end,zhu2021deformable}, segmentation~\cite{chen2021pre}, etc. ViT~\cite{dosovitskiy2021an} was the first attempt to introduce MHA (multi-head attention) architecture for image modality and surpassed the CNNs performance on image classification on large scale datasets. Later, DeiT~\cite{touvron2021training} further boost the performance of raw ViTs with the same architecture but with token-based knowledge distillation to enhance the representation learning. MAE~\cite{he2022masked} introduces a supervision technique to pretrain ViT encoder on masked image reconstruction pretext task and achieves state-of-the-art performance on ImageNet classification task. Swin Transformer~\cite{liu2021swin} utilized shifted window to introduce inter-window information exchange and enhance local attention. Transformer-iN-Transformer (TNT)~\cite{han2021transformer} aggregated both patch- and
pixel-level representations by a nested self-attention within each transformer block.

\subsection{Pruning on CNNs}

CNNs pruning has been widely studied for decades. Large amount of pruning methods can be categorized into unstructured pruning, semi-structured pruning, structured (channel/filter-wise) pruning, etc., depending on the the level of sparsity. 

\noindent\textbf{Unstructured Pruning}
removes individual connections (neurons) from convolution kernels, which is the earliest established pruning scheme by the pioneer works~\cite{han2015deep,han2015learning} that attempted unstructured pruning for LeNet and AlexNet. 
Following that, \cite{molchanov2016pruning,frankle2018the,zhu2018prune,morcos2019one} adopts magnitude-based or taylor-based criterion importance scores to threshold low-scored connections globally. 
\cite{gale2019state,evci2020rigging} leverage architectural heuristics to determine layerwise pruning rate.
Several efforts\cite{molchanov2016pruning,lee2019snip,lamp} consider the influence of pruning on the model output to determine layerwise pruning rate. 
\cite{isik2022information} assumed laplacian distribution of CNN weights to approximate output distortion. 
\cite{wang2022rdo,xu2023efficient} leverage rate-distortion theory to derive layer-wise pruning ratios that achieves optimal rate-distortion performance. 

\noindent\textbf{Structured Pruning}
or channel/filter-wise pruning scheme prunes the entire kernel in a Conv layer or a channel in fully connected layer at once. 
\cite{luo2017thinet} used feature map importance as a proxy to determine removable channels.
\cite{he2017channel} took a regularization based strucrtural pruning method.
\cite{yu2018nisp} obtains channel-wise importance scores by propagating the score on the final response layer. 
\cite{lin2020hrank} utilized rank informtation of feature maps to determine the prunable channels.
\cite{wang2022rdo} leveraged rate-distortion theory to prune the channels that lead to least model accuracy drop.
\cite{shen2022structural} took first-order importance on channels and allocates sparsities by solving a knapsack problem on all channel importances in the whole network.

\noindent\textbf{Semi-Structured Pruning}
is a less-explored approach that leverages sparsity pattern in between unstructured and structured pruning, where patterns such as block-sparsity in matmul can greatly benefit the realworld speedups by exploiting the nature of GPU calculation~\cite{mao2021tprune,lagunas2021block}. With the sparsity pattern less agressive than sturctured pruning, the impact of removing neurons on the model accuracy is less than structured pruning. Nevertheless, semi-structured pruning is under-explored on the emerging ViTs, which are constructured with transformer encoder architecture with mostly fully connected layers.

\subsection{Sparsity in ViTs}
Witnessing the success of CNNs pruning, ViTs pruning is also receiving emerging interests. Compared to CNNs pruning, less efforts are devoted to pure weight pruning but more on pruning of tokens, MHA, etc.
S$^2$ViTE~\cite{chen2021chasing} first proposed to prune out tokens as well as self-attention heads under structured pruning scheme with sparse training for ViTs.
UVC~\cite{yu2021unified} derived a hybrid optimization target that unifies structural pruning for ViT weights, tokens and skip configuration to achieve sparse training for ViTs.
SPViT~\cite{kong2022spvit} only performed token pruning on attention heads but adopted latency constraint to maximize speedup on edge devices. 
\cite{yang2023global} adopts Nvidia's Ampere 2:4 sparsity structure to achieve high speedup but required structural constraints to ensure a matching dimensions of qkv, feedforward and projection layers (head alignment) to search for subnetwork from larger ViT variants to match the latency of smaller ones. Unlike prior works~\cite{uvc,yang2023global}, our method focuses on pure weight pruning scheme and does not require heavy searching for the coordination of different compression schemes. 

Some efforts~\cite{kitaev2019reformer,wu2019lite,wang2021spatten,zaheer2020big} sparsify the heavy self-attention by introducing sparse and local attention patterns for language models. \cite{child2019generating} attempts on ViTs, but these sparse attention schemes still require training from scratch.

\section{Methodologies}

\subsection{Preliminaries}
\noindent\textbf{Block-structured Pruning within layer.}
We targeted at block-structured pruning for all linear layer weights, which include any parametrized linear layers in the ViTs, such as qkv layers, feedforword and projection layers. Neurons in these weight matrices are grouped in 2-dimensional fixed-sized blocks as a unit for pruning. 
To decide which blocks need to be pruned, given a block structure $(B_h, B_w)$, for each matrix $\bm{W}\in \sR^{H\times W}$, we rank the blocks by the average of 1st order taylor expansion score of the neuron within each block. Mathematically, we first obtain the neuron score by the taylor expansion $\mS = |\mW\cdot\nabla_{\mW} f|$ similar to~\cite{molchanov2019importance}, then perform a 2D average pooling to obtain a score for each block $\mS^\prime\in\sR_\ast^{H/B_h \times W/B_w}$ ($\sR_\ast$ is non-negative real value set). Then given a pruning ratio for each layer, we can rank the blocks by their scores and eliminate the bottom ranked ones. The right most part of \cref{fig:blocksparse} visualizes the block-structure patterns realistically generated from ViTs. 
The above pruning scheme can be formulated as $\widetilde{\mW}_{i,j}=\mW_{i,j}\odot \mM_\alpha(\mS^\prime)_{\ceil{\frac{i}{B_h}},\ceil{\frac{j}{B_w}}}$, where $\mM_\alpha(\mS^\prime)$ is the binary mask generated from the previous block-wise score matrix under the pruning ratio $\alpha$.

\noindent\textbf{Pruning scheme of ViTs.}
Unlike prior arts, the scope of this work is only eliminating model parameters to reduce computation, without considering 
other aspects of ViTs like token number and token size and transformer block skipping~\cite{chen2021chasing,yu2021unified,kong2022spvit}.

We further adopt a basic assumption for the weight perturbation $\Delta\mW = \widetilde{\mW} - \mW$ caused by a typical pruning operation to the weight:
\begin{assumption}\label{assum:ind-zeromean} 
    \textbf{I.i.d. weight perturbation across layers}~\cite{zhou2018adaptive}: This means the joint distribution is zero-meaned: 
    \begin{equation}
        \forall{0<i\ne j<L}, E(\Delta\mW^{(i)}\Delta\mW^{(j)}) = E(\Delta \mW^{(i)}) E(\Delta \mW^{(j)}) = 0,
    \end{equation}
    and also zero co-variance: $E(\|\Delta\mW^{(i)}\Delta\mW^{(j)}\|^2)=0$.
\end{assumption}

\begin{figure}[t]
    \centering
    \includegraphics[width=\textwidth]{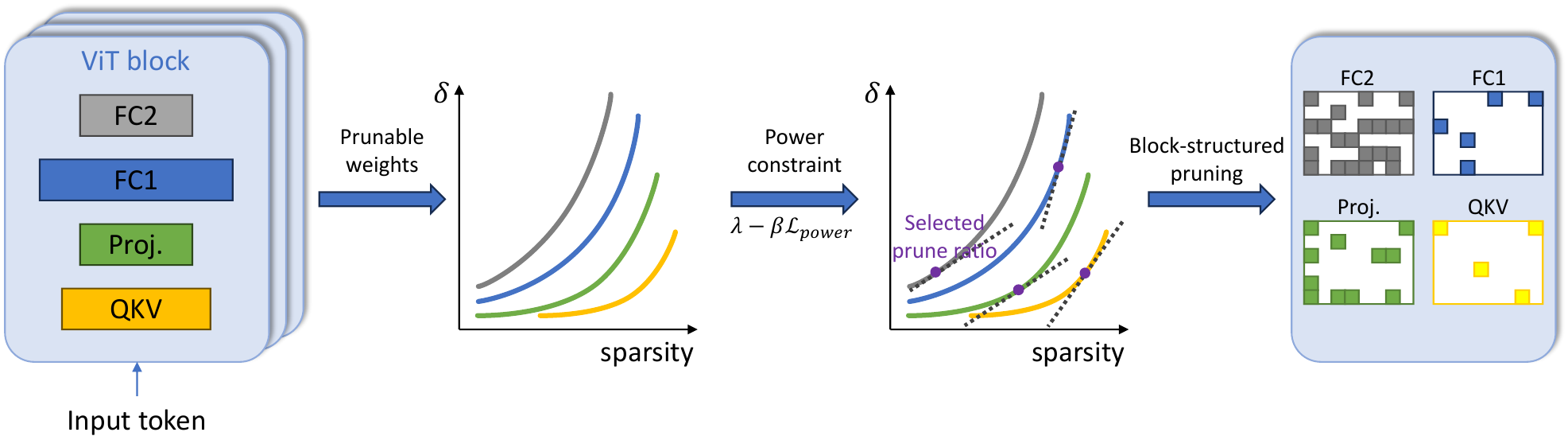}
    \caption{Illustration of the proposed Low Power Semi-structured pruning method. Widths of different layers within ViT block visualizes the computation complexities (FLOPs) of single layer. We first extract all layers with prunable weights in the pretrained ViTs, then we obtain the empirical curves $\delta$-vs-sparsity as described in Eq.~\ref{eq:langrangian2}. We further calculate the layer specific target slope $\lambda_i$ according to its contribution to the power consumption and select the layer-wise pruning ratios when the target slopes are tangential to the curves. Finally we prune the layer weights given their pruning ratios in block-structured sparsity, and finally finetune the pruned ViTs. The rightmost of the diagram is an example of the block-sparsity patterns when block sizes for both dimensions are the same, but they don't have to be the same as in the experiment section.}
    \label{fig:blocksparse}
    \vspace{-20pt}
\end{figure}

\subsection{Hardware-aware pruning objective}
Since layers may contribute differently to the model performance~\cite{frankle2020pruning}, various criteria have been proposed to allocate layerwise sparsity given a total budget~\cite{lee2019snip,isik2022information,lamp,xu2023efficient}. 
However, most existing pruning objectives can be summarized as minimizing the model output accuracy under computation constraint, without explicitly taking into account the actual power consumption and speedup. 
In contrast, our compression pruning objective directly optimizes the power consumption to achieve certain computation reduction target (FLOPs).
Specifically, given a neural network $f$ of $l$ layers and its parameter set $\mW^{(1:l)} = \big(\mW^{(1)},...,\mW^{(l)}\big)$, where $\mW^{(i)}$ is the weights in layer $i$, pruning parameters in the $f$ will give a new parameter set $\widetilde{\mW}^{(1:l)}$.
We view the impact of pruning as the distance between the network outputs $f(x;\mW^{(1:l)})$ and $f(x;\widetilde{\mW}^{(1:l)})$.

Hence our learning objective is as follows: 
\vspace{-8pt}
\begin{equation} \label{eq:obj}
\begin{split}
 \min &\ \|f(x;\mW^{(1:l)}) - f(x;\widetilde{\mW}^{(1:l)})\|^{2} + \beta \gL_{power}(f(\widetilde{\mW}^{(1:l)})), \\ s.t. &\ \frac{\mathrm{FLOPs}(f(\widetilde{\mW}^{(1:l)})}{\mathrm{FLOPs}(f(\mW^{(1:l)})} \leq R,
 \end{split}
\vspace{-8pt}
\end{equation}
which jointly minimize the output distortion caused by pruning (first term) as well as the estimated power consumption $\gL_{power}(f(\widetilde{\mW}^{(1:l)}))$, under a certain FLOPs reduction target $R$. 

\subsection{Second-order Approximation of Output Distortion}
To solve the pruning objective, we break down the first term related to the output distortion. 
We first expand the output distortion $f(x;\mW^{(1:l)}) - f(x;\widetilde{\mW}^{(1:l)})$ using second-order taylor expansion: (omit the superscript $(1:l)$ for visual clarity from now)
\vspace{-5pt}
\begin{equation}\label{eq:second_taylor}
    f(x;\mW) - f(x;\widetilde{\mW}) = \sum_{i=1}^{l}{\nabla_{\mW^{(i)}}^\top f \Delta\mW^{(i)} + \frac{1}{2} \Delta\mW^{(i)\top} \mH_i\Delta\mW^{(i)}},
\vspace{-5pt}
\end{equation}
where $\mH_i$ is the hessian matrix of the $i$-th layer weight.

Then consider the expectation of the squared L2 norm in the objective~Eq.~\ref{eq:obj}, which can be rewritten as the vector inner-product form:
\begin{equation}\label{eq:expl2norm}
\resizebox{\textwidth}{!}{$
\begin{split}
    &E(\|f(x;\mW) - f(x;\widetilde{\mW})\|^2) = E\left[(f(x;\mW) - f(x;\widetilde{\mW})^\top(f(x;\mW) - f(x;\widetilde{\mW})\right] \\
    &= \sum_{i,j=1}^l{E\left[\left(\nabla_{\mW^{(i)}}^\top f \Delta\mW^{(i)} + \frac{1}{2} \Delta\mW^{(i)\top}\mH_i\Delta\mW^{(i)}\right)^\top\left(\nabla_{\mW^{(j)}}^\top f \Delta\mW^{(j)} + \frac{1}{2} \Delta\mW^{(j)\top} \mH_j\Delta\mW^{(j)}\right)\right]}.
\end{split}$}
\end{equation}
When we further expand the inner-product term, the cross-term for each pair of different layer $1\le i\neq j \le l$ is:
\begin{equation}\label{eq:crossterm}
\resizebox{\textwidth}{!}{$
\begin{split}
    &E\left[\Delta\mW^{(i)\top}\nabla_{\mW^{(i)}}f\nabla_{\mW^{(j)}}^\top f\Delta\mW^{(j)}\right] + E\left[\frac{1}{2}\Delta\mW^{(i)}\Delta\mW^{(i)\top}\mH_i^\top\nabla_{\mW^{(j)}}^\top f\Delta\mW^{(j)} \right] + \\
    &E\left[\frac{1}{2}\Delta\mW^{(i)\top}\nabla_{\mW^{(i)}}f \Delta\mW^{(j)\top} \mH_j\Delta\mW^{(j)}\right] + E\left[\frac{1}{4}\Delta\mW^{(i)}\Delta\mW^{(i)\top}\mH_i^\top \Delta\mW^{(j)\top} \mH_j\Delta\mW^{(j)}\right].
\end{split}$}
\end{equation}
When we dicuss the influence of the random variable $\Delta \mW$, the first-order and second-order derivatives $\nabla_{\mW}f$ and $\mH$ can be regarded as constants and therefore can be moved out of expectation. Also vector transpose is agnostic inside expectation. So Eq.~\ref{eq:crossterm} becomes
\begin{equation}\label{eq:crossterm2}
\begin{split}
    &\nabla_{\mW^{(i)}}f\nabla_{\mW^{(j)}}^\top f E(\Delta\mW^{(i)\top}\Delta\mW^{(j)}) + \frac{1}{2}\mH_i^\top\nabla_{\mW^{(j)}}^\top f E(\Delta\mW^{(i)}\Delta\mW^{(i)\top}\Delta\mW^{(j)}) +\\
    &\frac{1}{2}\nabla_{\mW^{(i)}}f \mH_jE(\Delta\mW^{(i)\top}\Delta\mW^{(j)\top}\Delta\mW^{(j)}) + \frac{1}{4}\mH_i^\top\mH_j E(\|\Delta\mW^{(i)\top}\Delta\mW^{(j)}\|^2).
\end{split}
\end{equation}
Using~\cref{assum:ind-zeromean}, we can find that the above 4 cross-terms also equal to zero~\footnote{We empirically find $E(\mW^{(i)\top} \mW^{(i)} \mW^{(j)})=0$ holds on top of $E(\mW^{(i)} \mW^{(j)})=0$.}. Therefore the expectation Eq.~\ref{eq:expl2norm} results in only intra-layer terms:
\begin{equation}\label{eq:additiv}
    E(\|f(x;\mW) - f(x;\widetilde{\mW})\|^2)= \sum_{i=1}^l{E\left(\left\|\nabla_{\mW^{(i)}}^\top f \Delta\mW^{(i)} + \frac{1}{2} \Delta\mW^{(i)\top}\mH_i\Delta\mW^{(i)}\right\|^2\right)}.
\end{equation}

\subsection{Power consumption under Block-structured Pruning}
As the majority of the power consumption of network inference is attributed to the matrix multiplication operation, the network power consumption can be estimated by summing individual power cost of block-sparse matrix multiplication of each linear layers. Consider a matrix $\mA\in\sR^{M\times N}$, typically input tensor, to be multiplied with the block-sparse weight matrix $\mB\in\sR^{N\times K}$ with block-structure of $(B_n, B_k)$ and $\alpha$-percentage of blocks pruned out. When using a block-sparse GEMM configured with the kernel grid size of $B_m$ on $M$-dimension, the power consumption of the block-sparse matmul can be estimated as 
\vspace{-10pt}
\begin{equation}
    P = p_m \frac{M}{B_m}\ceil{(1-\alpha)\frac{N}{B_n}\frac{K}{B_k}},
\vspace{-10pt}
\end{equation} 
where $p_m$ is the power cost of individual within-block matmul. Therefore, the second term in Eq.~\ref{eq:obj} can be obtained by adding up the power consumption of the network of all layers:
\vspace{-10pt}
\begin{equation}\label{eq:power_raw}
    \beta\gL_{power} = \beta p_m \sum_{i=1}^l{\frac{M_i}{B_m}\ceil{(1-\alpha_i)\frac{N_i}{B_n}\frac{K_i}{B_k}}},
\vspace{-10pt}
\end{equation}
where $p_m$ and $B_m$ can be absorbed into the weight coefficient $\beta$ because they only depends on hardware parameters and GEMM configuration which is unified across layers. 

\noindent\textbf{Final Objective.} Combining~Eq.~\ref{eq:additiv} and Eq.~\ref{eq:power_raw}, the final objective can be reformulated as:
\begin{equation}\label{eq:obj_f}
\vspace{-10pt}
\resizebox{\textwidth}{!}{$
\begin{split}
 \min &\ \sum_{i=1}^l{E\left(\left\|\nabla_{\mW^{(i)}}^\top f \Delta\mW^{(i)} + \frac{1}{2} \Delta\mW^{(i)\top}\mH_i\Delta\mW^{(i)}\right\|^2\right)} + \beta \sum_{i=1}^l {M_i\ceil{(1-\alpha_i)\frac{N_i}{B_n}\frac{K_i}{B_k}}} \\
 s.t.&\ \frac{\mathrm{FLOPs}(f(\widetilde{\mW}^{(1:l)})}{\mathrm{FLOPs}(f(\mW^{(1:l)})} \leq R.
\end{split}
$}
\vspace{-10pt}
\end{equation}

\subsection{Finding Solution to Pruning Objective}\label{sec:solution}
At this point, we can further solve the optimization problem Eq.~\ref{eq:obj_f} on the layer-wise pruning ratio set $\{\alpha_i \mid 1\le i \le l \}$ by applying lagrangian formulation~\cite{wang2022rdo,xu2023efficient}
\vspace{-10pt}
\begin{equation}\label{eq:lagrangian}
    \frac{\partial}{\partial \alpha_i}\left( \left\|\nabla_{\mW^{(i)}}^\top f \Delta\mW^{(i)} + \frac{1}{2} \Delta\mW^{(i)\top}\mH_i\Delta\mW^{(i)}\right\|^2 + \beta M_i\ceil{(1-\alpha_i)\frac{N_i}{B_n}\frac{K_i}{B_k}}\right) = \lambda.
\vspace{-10pt}
\end{equation}
In practice we can get rid of the ceiling function in Eq.~\ref{eq:lagrangian} and therfore:
\vspace{-10pt}
\begin{equation}\label{eq:langrangian2}
    \frac{\partial}{\partial \alpha_i}\left(\left\|\nabla_{\mW^{(i)}}^\top f \Delta\mW^{(i)} + \frac{1}{2} \Delta\mW^{(i)\top}\mH_i\Delta\mW^{(i)}\right\|^2\right) = \lambda_i = \lambda + \beta\frac{M_i N_i K_i}{B_n B_k},
\vspace{-10pt}
\end{equation}
which will give a continuous $\alpha_i\in[0,1]$ compared to the original solution with the ceiling, but in practice since the number of blocks within a weight tensor is limited the pruning ratio $\alpha_i$ is to be rounded to a discrete value anyway.  
Solving Eq.~\ref{eq:langrangian2} will need to collect empirical curves for all layers (pruning ratio $\alpha_i$ against the taylor second-order term $\delta_i=\nabla_{\mW^{(i)}}^\top f \Delta\mW^{(i)} + \frac{1}{2} \Delta\mW^{(i)\top}\mH_i\Delta\mW^{(i)}$). 
By setting a specific $\lambda$, we can solve Eq.~\ref{eq:langrangian2} individually for each layer by searching for a $\alpha_i$ that let the equality holds. 
The final solution of pruning ratios can be obtained by traversing $\lambda$ that returns a pruned network closest to the constraint $R$.

One \emph{key insight} that one can derive from the optimization solution Eq.~\ref{eq:langrangian2} is that by controlling the weight $\beta$, the power consumption are explicitly incorporated in the optimization process in the form of altering the target slope for the partial derivative of the curve $\frac{\partial \delta_i(\alpha_k)}{\partial \alpha_k}$, which represents how intensely pruning one layer affects the final model accuracy (output distortion). In this way, we achieve direct tradeoff between model accuracy and power consumption.

\subsection{Empirical Complexity}
\noindent\textbf{Hessian approximation.} For empirical networks, we approximate the hessian matrix $\mH_i$ using \emph{empirical fischer}~\cite{kurtic2022optimal}:
\vspace{-10pt}
\begin{equation}
    \mH_i = \mH_{\gL}(\mW^{(i)}) \approx \hat{\mF}(\mW^{(i)}) = \kappa \mI_d + \frac{1}{N}\sum_{n=1}^N {\nabla_{\mW^{(i)}} f_n \nabla_{\mW^{(i)}}^\top f_n}.
\vspace{-10pt}
\end{equation}
In order to obtain empirical curves $\frac{\partial \delta_i(\alpha_k)}{\partial \alpha_k}$ on a calibration set, one is possible to traverse different pruning ratio (\emph{e.g.} in practice $\alpha_k = \frac{k+1}{K}, 0<k<K$) and caluculate the corresponding $\delta_i (\alpha_k)$ for all $0<k<K$. However in such case, even with the approximated hessian, the curve generation for each layer is still very expensive at the complexity of $O(NKD_i^4)$, where $K$ is the number of possible pruning ratio selections and $D_i=N_i K_i$ is the dimension of weight in $i$-th layer. This poses challenge to make the proposed method efficient enough to enjoy the benefits of sparse network. 
We notice that the derivative $\nabla_{\mW_i}$ is constant to the change of pruning ratio which let us to reuse the hessian matrix for all pruning ratio, which drops the complexity to $O((N+K)D_i^2 + KD_i^4)$. 
However, the existence of the biquadratic complexity makes it still too expensive. 
We further notice that when pruning ratio move up slightly, only a partition of the weight vector is pruned out from $\widetilde{\mW_i}$. Therefore we can select a subvector $\mathrm{d}\Delta\mW_i(\alpha_k) = \Delta\mW_i(\alpha_{k}) - \Delta\mW_i(\alpha_{k-1})$ each time when pruning ratio increases from $\alpha_{k-1}$ to $\alpha_{k}$ and update the $\delta_i(\alpha_k)$ from $\delta_i(\alpha_{k-1})$ by the following rule:
\vspace{-12pt}
\begin{equation}\label{eq:update}
\resizebox{\textwidth}{!}{$
\begin{split}
    \delta_i(\alpha_k) - \delta_i(\alpha_{k-1}) &= \nabla_{\mW^{(i)}}^{\top\prime} f \mathrm{d}\Delta\mW_i(\alpha_{k}) + \left(\frac{1}{2}\mathrm{d}\Delta\mW_i(\alpha_{k}) + \Delta\mW_i(\alpha_{k-1})\right)^\top\mH_i^\prime \mathrm{d}\Delta\mW_i(\alpha_{k}).
\end{split}$}
\vspace{-12pt}\end{equation}
Denote the dimension of the subvector $\mathrm{d}\Delta\mW_i(\alpha_k)$ as $d_i(k) \ll D_i$ equals the number of values changes from $\Delta\mW_i(\alpha_{k-1})$ to $\Delta\mW_i(\alpha_{k})$, the multiplication calculation in Eq.~\ref{eq:update} can be operated at lower dimensions, where $\nabla_{\mW^{(i)}}^{\top\prime} f\in\sR^{d_i(k)}, \mH_i^\prime\in\sR^{D_i\times d_i(k)}$ are subvector and submatrix indexed from the original ones. At $k=1$, $\alpha_k=0$ \emph{i.e.} there is no pruning at all which guarantees $\delta_i(\alpha_1)=0$.
Since $\alpha_k$ increases linearly, the $d_i(k)\approx \frac{D_i}{K}$, therefore, the complexity is around $O(\frac{N}{2}D_i^2)$.
Hence the total computation complexity for calculating $\delta$ for all $l$ layers is around $O(\frac{1}{2}\sum_i^l{D_i^2})$, which is far less than the original complexity. 

To this end, we presented a hardware-aware pruning criterion that explicitly accounts for the power consumption of the block-structured sparse model inference. The block-structured pruning scheme enables the obtained sparse network to achieve real-world acceleration on hardware while optimally preserving the network accuracy. 
The algorithm is extremely efficient to obtain a sparse ViT. 

\section{Experiments}

\subsection{Datasets and Benchmarks}
We extensively evaluate our ViTs pruning method for image classification task as well as a downstream image segmentation task.
For classification task, we conduct experiments mainly on Deit~\cite{touvron2021training} model as well as Swin Transformers~\cite{liu2021swin} on ImageNet dataset~\cite{krizhevsky2012imagenet}. We adopt the same training settings as in UVC~\cite{uvc} for the finetuning of ViTs. 
For image segmentation, we test on the transferred performance of DeiT-Base backbone in SETR~\cite{zheng2021rethinking} model on Cityscapes dataset~\cite{Cordts2016Cityscapes}.
More implementation details can be found in supplementary materials.

\noindent\textbf{Baseline methods.}
For the following experiments, we followed the UVC~\cite{uvc} comparison settings and compare ourselves to the previous ViTs compression methods that at least involves model weights pruning, as well as hybrid methods, including SCOP~\cite{tang2020scientific}, VTP~\cite{zhu2021vision}, S$^2$ViTE~\cite{chen2021chasing}, X-Pruner~\cite{yu2023x} and UVC~\cite{uvc} itself. We also include a uniform pruning result for Deit-Base where it fixes a uniform pruning ratio for all layers.

\subsection{Main results}

\begin{table}[t]
\caption{Comparisons with state-of-the-art ViTs pruning methods. We label the sparsity schemes the baselines exploit for a fair comparison, where \textbf{S}: Structured Sparsity, \textbf{U}: Unstructured Sparsity, \textbf{B}: Blocksparse, \textbf{T}: Token selection(gating), \textbf{A}: Attention. For LPViT (Ours), we report the best results among different blocksize configurations.}\label{tab:main-result}\vspace{-10pt}
\centering
\begin{tabular}{cccccc}
\hline\hline
Model                       & Method     & \makecell{Sparsity\\Scheme}  & FLOPs(G) & \makecell{FLOPs remained (\%)} & \makecell{Top-1 Acc (\%)} \\ \hline
\multirow{6}{*}{Deit-Small} & Dense      &    & 4.6      & 100   &  79.8                    \\ 
                            & Uniform    & \textbf{U}  & 2.3      & 50         & 77.17 (-2.63)           \\
                            & SCOP       & \textbf{S}  & 2.6      & 56.4       & 77.5 (-2.3)            \\ 
                            & S$^2$ViTE  & \textbf{S+T}  & 3.14     & 68.36      & 79.22 (-0.58)          \\ 
                            & UVC        & \textbf{S+A+T}  & 2.32     & 50.41      & 78.82 (-0.98)          \\ 
                            & \textbf{LPViT} & \textbf{B}  & 2.3      & 50         & $\mathbf{80.69 (+0.89)}$          \\\hline
\multirow{7}{*}{Deit-Base}  & Dense     &     & 17.6     & 100   & 81.8                      \\ 
                            & S$^2$ViTE  & \textbf{S+T}   & 11.87    & 66.87& 82.22 (+0.42)               \\ 
                            & VTP        & \textbf{U}    & 10       & 56.8   & 80.7 (-1.1)              \\ 
                            & UVC        & \textbf{S+A+T}  & 8        & 45.5   & 80.57 (-1.23)              \\ 
                            & \textbf{LPViT} & \textbf{B}  & 8.8     & 50   & $\mathbf{80.81 (-0.99)}$      \\ 
                            & \textbf{LPViT} & \textbf{B}         & 7.92     & 45   & 80.55 (-1.25)   \\           & \textbf{LPViT} & \textbf{B+A+T}  & 5.10 & 29   & $\mathbf{81.03 (-0.77)}$\\ \hline
\multirow{2}{*}{Swin-Base}  & Dense &  & 15.4 & 100 &  83.5 \\
        & \textbf{LPViT} & \textbf{B}  & 11.24 & 73 & 81.73 (-1.77) \\\hline
\multirow{5}{*}{Swin-Tiny}  & Dense &  & 4.5 & 100 & 81.2 \\
        & X-Pruner & \textbf{S+A}  & 3.19 & 71.1 & 78.55 (-2.65) \\
        & \textbf{LPViT} & \textbf{B} & 3.47 & 77 & $80.0 (-1.2)$  \\
        & \textbf{LPViT} & \textbf{B}  & 3.19 & 71.1 &  $\mathbf{79.24 (-1.96)}$   \\ 
        & \textbf{LPViT} & \textbf{B+A}  & 2.72 & 60.47 &  $\mathbf{80.0 (-1.2)}$ \\\hline\hline
\end{tabular}\vspace{-10pt}
\end{table}

As presented in Tab.~\ref{tab:main-result}, we first notice that our result on Deit-Small achieves loss-less, and even higher than dense model performance by $0.89$, with roughly the same FLOPs, surpassing existing baselines by a \textbf{large margin}. 
On larger architectures like DeiT-Base, where our method displays less prominent improvement but still on-par performance on the Top-1 accuracy of $80.81$ with $50\%$ FLOPs remaining and $80.55$ with around $45\%$ FLOPs. 
This is an intuitive observation since coarser pruning patterns like structural pruning would hurt the performance of smaller models more than larger model with a lot more redundant weights, and that is also where smaller structures such as the proposed block-sparsity pattern will retain more performance while still ensure speedup compared to unstructured pruning.
On DeiT-Small, we beat all existing hybrid methods that leverage patch-slimming or token selections. We remain competitive on larger model DeiT-Base, while we notice S$^2$ViTE cannot achieve comparable FLOPs reduction to us.
On non-global attention transformers such as Swin Transformer, LPViT remains competitive compared to other ViT pruning methods, achieving only $1.96\%$ loss on Swin-Tiny with FLOPs $71.7\%$. We also report a mere $1.77$ drop on Swin-Base.
We show less improvement on DeiT-B despite more parameters than DeiT-S. 
Although seems to be counterintuitive, the reason can be that larger models are easier to get overfitted at finetuning stage so that accuracy cannot be fully recovered.
This can also be attributed to the regularizing effects of pruning, benefitting to the OOB data performance, and has been discovered in many prior papers~\cite{liu2023pd}.
Meanwhile, LPViT is a generic framework which also supports integration with other sparsity.
When combined with attention and token pruning, LPViT further improves accuracy with even less FLOPs for DeiT-Base and Swin-Tiny.

\begin{figure}[t]
    \centering
    \includegraphics[width=\linewidth]{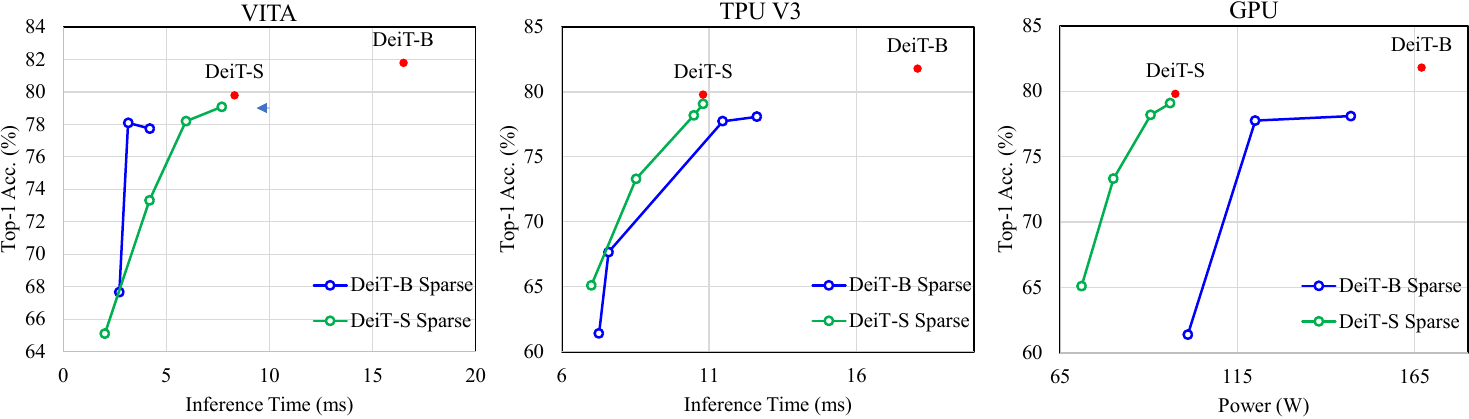}
    \caption{Inference overhead and power reductions on hardware platforms. }
    \label{fig:hardware_curves}
\end{figure}

\begin{table}[t]
    \centering
    \caption{Frontier speedup Results on various hardware platforms and Power Reduction on A-100 GPU.}
    \label{tab:hardware}
    \begin{tabular}{c|ccc|ccc|c}\hline\hline
        Model & \makecell{FLOPs\\ Remained (\%)} & Blocksize  & \makecell{Top-1\\Accuracy} & VITA & TPU-V3 & A-100 & Power (W) \\ \hline
        \multirow{3}{*}{DeiT-Base} & $100$ & -  & $81.8$ & - & - & - & $167$ \\ 
         & $25.8$ & $32\times32$ & $77.75$ &  $\mathbf{3.93}\times$ & $1.57\times$ & $1.75\times$ & $\mathbf{120 (71.8\%)}$  \\ 
         & $38.5$ & $64\times32$ & $78.1$ & $2.61\times$ & $1.43\times$ & $1.79\times$ & $147 (88\%)$  \\ \hline
         \multirow{3}{*}{DeiT-Small} &  $100$ & - & $79.8$ & -& - & -   & $195$ \\ 
         & $71.3$ & $16\times 32$ & $78.2$ & $\mathbf{1.4}\times$ & $1.03\times$ & $1.2\times$ & $\mathbf{181 (92.8\%)}$ \\ 
         & $75$ & $16\times 16$ & $80.65$ & $1.33\times$ & $1.09\times$ & $1.14\times$ & / \\\hline\hline
    \end{tabular}\vspace{-10pt}
\end{table}

\subsection{Hardware Performance Benchmarks}
We evaluate the inference efficiencies of our pruned models in terms of speedup and power consumption reductions on three types of hardware platforms and summarize the results in \cref{tab:hardware}. We first simulated the statistics of ViTs on a RISC-V platform ViTA~\cite{chen2024vita}, which supports GEMMs in various blocksizes, enabling us to fulfill the theoretical speedups of blocksparse models. 
Our approach thereby obtains as high as $\mathbf{3.93}\times$ and $\mathbf{1.4}\times$ speedups for DeiT-Base and DeiT-Small respectively.
We then perform simulation on a high-bandwidth platform Google TPU-V3~\cite{jouppi2020domain}. Since TPU V3 only offers 1 type of MAC with block size fixed at $128\times128$, we expect less speedup than on RISC-V. Nevertheless, we still see positive $\mathbf{1.57}\times$ speedup for DeiT-Base. 
Finally, we deploy on NVIDIA A100 40GB GPU with CUDA 11.8 and evaluate the end-to-end inference time and the runtime power consumption, and observe a power reduction up to $71.8\%$ on DeiT-Base. 
Power consumption is measured by averaging \texttt{nvidia-smi}’s power meter over an adequate time period and subtracting the idle power consumption.

\cref{fig:hardware_curves} shows the results when pruning the models to different sizes, \textit{i.e.} the inference time and accuracy at different pruning levels.
Specifically, each dot represents one result for a specific pruning configuration. 
In \cref{fig:hardware_curves}, we observed that on VITA, DeiT-B obtains better trade-off than DeiT-S, while on TPU and GPU, DeiT-S performs better.
This shows that the models may perform differently on different hardware platforms.


\begin{table}[t]
    \caption{Ablation studies on the Power consumption constraint on the pruning result. We compare between the results with the power constraint (main results) and without (by setting $\beta=0$).}
    \label{tab:power}
    \centering
    \begin{tabular}{cccc}
    \hline\hline
        Method                                      & Top-1 Acc (\%)    & Params remained (\%) & FLOPs remained (\%)       \\\hline
        \multicolumn{4}{c}{Deit-Base-BK32BN32}                                                            \\\hline
        w/ Power constraint                                & 80.81             & 73.3 & 52.5                                    \\
        w/o Power constraint                            & 77.75              & 26.9  & 55.6                      \\\hline
        \multicolumn{4}{c}{Deit-Base-BK32BN64}                                                            \\\hline
        w/ Power constraint                                 & 80.71             & 72.8 & 50                    \\
        w/o Power constraint                       & 61.42             & 49   & 49.7                    \\\hline\hline
\end{tabular}\vspace{-10pt}
\end{table}

\subsection{Ablation Study and Discussions}
Beyond the main results, we also attempt to discover how each creative parts in our proposed pruning scheme contribute to the final results, \emph{e.g.} the essential objective constraint regulating the power consumption and the block-sparsity structure, and answering the important questions such as why does the power constraint benefits the performance. We present the detailed ablation studies in the Tab.~\ref{tab:power} and Tab.~\ref{tab:abl}.

\noindent\textbf{Power constraint.}
We compared the behaviors of our pruning objective with and without the second-term power consumption in Eq.~\ref{eq:obj_f}.
As shown in Tab.~\ref{tab:power}, we notice that under the same level of FLOPs reduction rate of $50\%$ , our final pruning scheme (with power constraint) constantly gives significant higher results under different block shapes. 
This is an inspiring phenomenon since the power constraint are not designed to facilitate model accuracy at the first place. 
By inspecting the model sparsity, we learn that the proposed power constraint looks for layers with larger matmul dimensions to allocate more pruning quota to achieve most reduction in computation, and normally larger layers have more parameter redundancy. 
Therefore, this pruning ratio allocation actually cooperates with the main objective to minimize output distortion. 

\begin{table}[t]
\caption{Ablation study of different Block shape configurations on the pruning result.}\label{tab:abl}
\centering
\begin{tabular}{ccccc}
\hline\hline
Model                       & \makecell{Block shape\\(BK $\times$ BN)} & Sparsity (\%) & Top-1 Acc (\%) & \makecell{FLOPs\\remained (\%)} \\\hline
\multirow{4}{*}{Deit-Small} & $16\times16$                & 92.2          & 80.69            & 50                  \\
                            & $32\times16$                & 91            & 79.09            & 50                  \\
                            & $16\times32$                & 71.37         & 78.2             & 50                  \\
                            & $32\times32$                & 49            & 73.32            & 50                  \\\hline
\multirow{4}{*}{Deit-Base}  & $32\times32$                & 72.84         & 80.81            & 52.5                  \\
                            & $64\times32$                & 33.93         & 80.05            & 50                 \\
                            & $32\times64$                & 16.99         & 80.71            & 50                  \\
                            & $64\times64$                & 73.34         & 79.52            & 50.2                 \\\hline\hline
\end{tabular}
\vspace{-20pt}
\end{table}
\noindent\textbf{Block structure configurations.}
How well our optimization scheme adapts to different block size configurations is crucial to generalize on different hardware platforms with different levels of parallelism. Therefore, we conducted an ablation studies varying different block shapes combinations as listed in Tab.~\ref{tab:abl}.
Firstly, we observe that smaller block sizes preserve more model accuracy. 
Secondly, we notice that smaller networks are more sensitive to the change of block shapes, 
where different block sizes results in drastic change to the resulting number of parameters left in the networks.


\noindent\textbf{Layerwise-sparsity Allocation.}
We visualize the optimization results for each individual settings in Tab.~\ref{tab:abl} as below.
we notice some interesting observations. First, LPViT preserved more connections in the last transformer blocks including the classification head on both DeiT-B and DeiT-S on different pruning ratios, and the classification head is almost kept unpruned in all cases.
Second, on both DeiT-B and DeiT-S, we notice the projection layers after MHA in particular are always getting high pruning ratio, showing that projection layers in ViTs have more redundancies and effect the model performance the least. The above patterns are all automatically learned from our second-order pruning layer-wise sparsity allocation algorithm, showing the effectiveness of our method. 

\begin{minipage}[!t]{\textwidth}

\end{minipage}



\begin{figure}[t]
\centering
\begin{minipage}[!t]{.35\textwidth}
    \centering
    \includegraphics[width=\textwidth]{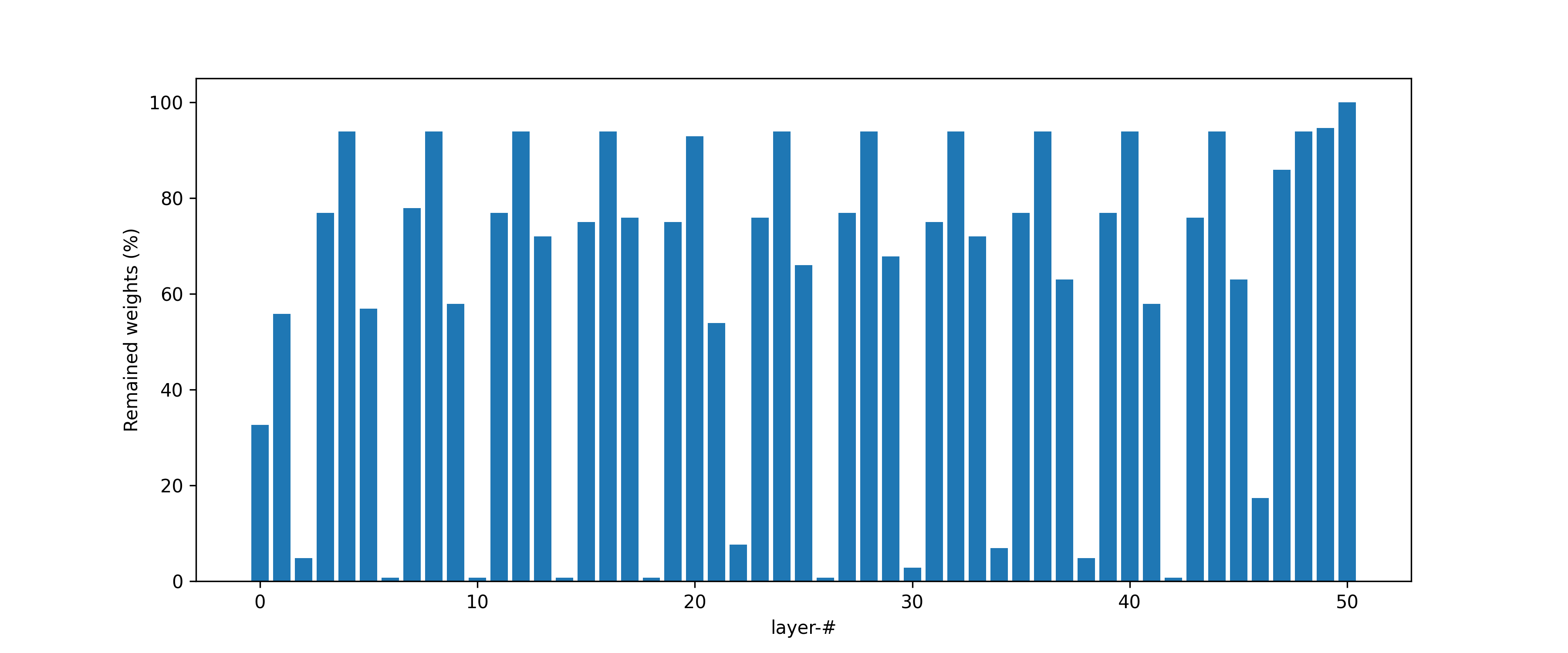}
    \makeatletter\def\@captype{figure}\makeatother\caption{Layerwise sparsity for DeiT-Base BK64BN64.}
    \label{fig:deit_base_bk64bn64_powerTrue_flops50}
\end{minipage}
\begin{minipage}[!t]{.6\textwidth}
    \centering
    \makeatletter\def\@captype{table}\makeatother\caption{Segmentation results on Cityscapes valiadtion dataset.}
    \label{tab:cityscapes}
    \begin{tabular}{cccc}
    \hline\hline
        Backbone & Method & FLOPs (G) & mIoU \\ \hline
        DeiT-Base/384 & Dense  & 17.6 & 78.66 \\
        DeiT-Base/384 & LPViT & 8.8 & 77.39 \\\hline\hline
    \end{tabular}
\end{minipage}
\vspace{-20pt}
\end{figure}

\noindent\textbf{Self-attention Maps. }
As shown in Fig.~\ref{fig:attn}, we followed the same self-attention maps visualization process adopted in ~\cite{chen2021chasing,cordonnier2019relationship} to show potential influence of the pruning on the attention behaviors in transformer multi-head attention. We observe that our LPViT block-sparsity pruning scheme displays a coarse and discretized pattern in a lot of attention heads across transformer blocks, accompanied by some completely inactive attention heads which allows us to further speedup inference by directly skiping those heads. 
Compared to structural pruning scheme, our semi-structured scheme allows middle states between blank attention and the delicate pattern in dense model, preserving more attention information which is curcial to the model accuracy. 
On LPViT-DeiT-Base (FLOPs 45\%), the last two attention layers have no active attention heads.
As a result, the entire blocks can be discarded in calculation, which could possibly bring the reported FLOPs reduction even more.

\noindent\textbf{Transfer Learning to Downstream Tasks.}
To evaluate the generalizability of LPViT on downstream tasks, we further evaluate on transferred learning performance of our method on the downstream Cityscapes~\cite{Cordts2016Cityscapes} segmentation task. We first pretrained a DeiT-B/384 pruned by 50\% FLOPS on imagenet for 27 epochs and further use it as a backbone in a recent segmentation model SETR~\cite{zheng2021rethinking} and train on Cityscapes dataset. Tab.~\ref{tab:cityscapes} compares the val mIoU of the pruned backbone and the original performance. 
We only observe a performance degration of mere 1.27 mIoU. 
More detailed results can be found in supplementary material.



\section{Conclusions}

In this work, we presented a novel ViTs weight pruning algorithm designed to reduce energy consumption during inference. Leveraging the linear layer-centric structure of the ViT architecture, we introduced a semi-structured pruning scheme to balance finetuning stability and hardware efficiency. Our algorithm is very efficient despite employing a hessian-based pruning criterion. Experimental results on various ViTs on ImageNet showcase the method's ability to identify optimal pruning solutions, maximizing accuracy for block-sparse models. Additionally, we illustrated the dual benefits of our proposed power-aware pruning objective, enhancing both software accuracy and hardware acceleration.

\section*{Acknowledgement}
This research is supported by the Agency for Science, Technology and Research (A*STAR) under its MTC Programmatic Funds (Grant No. M23L7b0021). Any opinions, findings and conclusions or recommendations expressed in this material are those of the author(s) and do not reflect the views of the A*STAR.

%
%
\bibliographystyle{splncs04}
\bibliography{output}
\end{document}

%% file: math_commands.tex
\usepackage{amsmath,amsfonts,bm}

\def\eqref#1{equation~\ref{#1}}

\def\ceil#1{\left\lceil #1 \right\rceil}

\def\1{\bm{1}}

\def\mA{{\bm{A}}}
\def\mB{{\bm{B}}}

\def\mF{{\bm{F}}}

\def\mH{{\bm{H}}}
\def\mI{{\bm{I}}}

\def\mM{{\bm{M}}}

\def\mS{{\bm{S}}}

\def\mW{{\bm{W}}}

\DeclareMathAlphabet{\mathsfit}{\encodingdefault}{\sfdefault}{m}{sl}
\SetMathAlphabet{\mathsfit}{bold}{\encodingdefault}{\sfdefault}{bx}{n}

\def\gL{{\mathcal{L}}}

\def\sR{{\mathbb{R}}}

